\documentstyle[latex-acl,fullname]{article}
\title{What is word sense disambiguation good for?}
\author{Adam Kilgarriff\thanks{\hspace{0.6em}Research supported by the EPSRC, grant
K18931.
Email: {\tt Adam.Kilgarriff@itri.bton.ac.uk}}\\
ITRI\\University of Brighton}

\begin{document}

\maketitle

\begin{abstract}
Word sense disambiguation has developed as a sub-area of natural
language processing, as if, like parsing, it was a well-defined
task which was a pre-requisite to a wide range of
language-understanding applications.  First, I review earlier work
which shows
that a set of senses for a word is only ever defined relative to a
particular human purpose, and that a view of word senses as part
of the linguistic furniture lacks theoretical underpinnings.   Then, I
investigate whether and how word sense ambiguity is in fact a problem for
NLP applications.
\end{abstract}

\section{What word senses are not}

There is now a substantial literature on the problem of word
sense disambiguation (WSD).  The goal of WSD research is generally
taken to be  disambiguation between the senses given
in a dictionary, thesaurus or similar.  The idea is simple
enough and could be stated as follows: 
\begin{quote}
 Many words have more than one meaning.  When a person understands a
sentence with an ambiguous word in it, that understanding is built on the
basis of just one of the meanings.  So, as some part of the human
language understanding process, the appropriate meaning has been chosen
from the range of possibilities.
\end{quote}
\noindent Stated in this way, it would seem
that WSD might be a well-defined task, undertaken by a particular
module within the human language processor.  This module could then be
modelled computationally in a WSD program, and this program,
performing, as it did, one of the essential functions of the human
language processor, would stand alongside a
parser as a crucial component of a broad range of NLP applications. 

There are problems with this view.  The simplest stems from the
observation that different dictionaries very often give different sets
of senses for a word.  A closer investigation reveals a lack of
theoretical foundations to the concept of `word sense'.  The concept
is intimately connected to our knowledge and experience of
dictionaries, but these are social artifacts created to satisfy such
human purposes as playing word-games, resolving family arguments, and
making profits for publishers.  Amid all these competing goals, the
pursuit of truth is not always dominant.

In particular, a standard dictionary specifies the range of
meaning of a word in a list, possibly nested, of senses.  This is not 
the outcome of an analysis of how word-meaning operates, but is,
rather, a response to constraints imposed by:
\begin{itemize}
\item tradition
\item the printed page
\item compactness
\item  a single, simple method of access
\item resolving disputes about what a word does and
does not mean.
\end{itemize}

The format of the dictionary has remained fairly
stable since Dr.\ Johnson's day.  The reasons for the format, and the
reasons it has proved so resistant to change and innovation, are
explored at length in \namecite{Nunberg:94}. In short, the development of
printed discourse, particularly the new periodicals, in England in the
early part of the eighteenth century brought about a re-evaluation of
the nature of meaning.  No longer could it be assumed that a
disagreement or confusion about a word's meaning could be settled
face-to-face, and it seemed at the time that the new discourse would
only be secure if there was some mutually acceptable authority on what
words meant.  The resolution to the crisis came in the form of
Johnson's Dictionary.  Thus, from its inception, the modern dictionary
has had a crucial symbolic role as in-principle arbiter of disputes. Hence ``the dictionary'', with its
implications of unique reference and authority (cf.\ ``the Bible'').  Further evidence for this position is to be found in
\namecite{McArthur:87}, for whom the ``religious or quasi-religious
tinge'' (p 38) to reference materials is an enduring theme in their
history; \namecite{Summers:88}, whose research into dictionary use found
that ``settl[ing] family arguments'' was a major use (p 114,
cited in  \namecite[p 151]{Bejoint:94}); and \namecite{Moon:89} who
catalogues the use of the UAD (Unidentified Authorising Dictionary)
from newspapers letters pages to restaurant advertising materials (pp
60--64).

To solve disputes about
meaning, a dictionary must be, above all, clear.  It must draw a line
around a meaning, so that a use can be classified as on one side of
the line or the other.  A dictionary which dwells on marginal or vague
uses of a word, or which presents word meaning as context-dependent or
variable or flexible, will be of little use for purposes of settling
arguments.  The pressure from this quarter is for the dictionary to
present a set of discrete, non-overlapping meanings for a word, each
defined by the necessary and sufficient conditions for its application
---whatever the facts of the word's usage.

Lexicographers are vividly aware of the problem.  They have
frequently lamented the possibly-nested list model
\namecite{Stock:83,Hanks:94,FillmoreAtkins:92}.  They know all too well
the injustice it frequently does to a word's range of meaning and
use. But WSD researchers, at least until recently, have generally
proceeded as if this was not the case: as if a single program
---disambiguating, perhaps, in its English-language version, between
the senses given in some hybrid descendant of Merriam-Webster, LDOCE,
COMLEX, Roget, OALDCE and WordNet ---would be relevant to a wide range
of NLP applications.\footnote{The most promising recent WSD work is
moving away from this position, determining the senses between which
the program is to disambiguate either directly from the clusters in
the corpus
\cite{Schutze:97}, or through a small amount of human input 
\cite{Clear:94}, or a choice of either \cite{Yarowsky:95}.}

The sets of word senses presented in different
dictionaries and thesauri have been prepared, for various purposes,
for various human users: there is no {\em a priori} reason to believe those sets are
appropriate for any NLP application.\footnote{For a full account
of the nature of word senses, in dictionaries and elsewhere, see
Kilgarriff (1992; 1993; 1997b).}

\nocite{thesis,ak:chum,believe}

It seems likely that NLP application lexicons ---which are, in the mid
1990s, almost invariably hand-built rather than MRD-derived--- will be
application-driven rather than resource-driven, so will only contain
the word senses and make the word sense distinctions relevant to the
application.  They might not encounter word sense ambiguity on
anything like the scale that a brief glance at a dictionary (or at the
WSD literature) would suggest.  The remainder of the paper addresses
whether this is so, and what scale of problem word sense ambiguity
causes for different varieties of NLP application.\footnote{My sources
include an informal email survey on the CORPORA mailing list,
to which I had 28 responses.}

\section{Taxonomy}

First, let us distinguish five
types of application for which WS ambiguity is potentially an issue:
\begin{itemize}
\item Information Retrieval (IR) 
\item Machine Translation (MT)
\item Parsing (and, implicitly, all those applications for which
parsing is one stage of processing)
\item Lexicography
\item Residual, `core' language understanding (including database front ends,
		dialogue systems, Information Extraction as in MUC)
		---hereafter NLU.
\end{itemize}

\subsection{IR}

The intellectual affinities of most recent WSD work are with IR. The
problem of finding whether a particular sense applies to an instance
of a word can be construed as equivalent to the essential IR task of
finding whether a document is relevant to a
query.  The homology is made explicit at various points in the
literature \cite{GCY:92,GCY:93}.

Most work in IR disregards syntactic
structure entirely, `stemming' words so that {\em clean, cleaner,
cleaning} and {\em cleaned} are all mapped to {\em clean}, and then
treats a document as a bag of stems.
It does not use POS-tagging or
name-recognition, although these are relatively mature and reliable
technologies for these tasks within NLP, and parsing has not 
been found to improve IR performance: the linguistic processing has not been fast, robust or
portable enough, and it is not in any case clear whether it
provides relevant information for the IR task.  This is very much a
live issue: see
\namecite{Strz:94}, \namecite{Strz:95} for recent evidence of the potential of NLP in
IR.  However, to date, IR has made progress through applying
sophisticated statistical techniques to documents treated as objects
without linguistic structure, and this is the approach to WSD which
has recently flourished.

  Within IR, WSD can be
viewed as an {\bf alternative} to NLP, rather than a technique within
it.  If a statistical model based on a bag of stems is inadequate, one
way to get closer to the meaning of a text is WSD; another is a
linguistically-informed technique such as parsing.  They are not
mutually exclusive, but nor are they readily compatible.

A high proportion of WSD research is oriented towards IR, yet it
is not clear whether WSD has the potential to significantly
improve IR performance.  In the first careful study of the question,
\namecite{KrovetzCroft:92} conducted some experiments which suggested
that WS-ambiguity causes only limited degradation
of IR performance.  Their experiments were on the
small, specialist CACM corpus.  They used a standard set of queries for which
``correct answers'' are available.  They compared system
performance `with ambiguity' and `without ambiguity': the `with
ambiguity' condition was the normal situation, while for the `without
ambiguity' condition, all relevant terms had been manually
disambiguated, in a simulation of a perfect WSD program.  For this
corpus and query-set, they concluded that a perfect WSD program would
improve performance by 2\%.

\namecite{Sanderson:94} performed a similar experiment using pseudo-words.
A pseudo-word is a word formed by `pretending' that
two distinct words were a single word with two meanings, one
corresponding to each of the original words.  Thus the pseudo-word
{\em banana-kalashnikov} could be formed by replacing all instances of
{\em banana} and {\em kalashnikov} in a corpus by {\em
banana-kalashnikov}: then a WSD program would have the task of
determining which were originally bananas, which kalashnikovs.  The
method allowed Sanderson to regulate the degree of ambiguity in the
corpus, and to model both accurate and inaccurate WSD programs.  He
found that introducing extra ambiguity did little to degrade
performance, but, when the WSD algorithm made mistakes, this did do
harm. Also, in longer queries the different words in the query will
tend to be mutually disambiguating, so WSD is probably only relevant
where the query is very short.  He concludes ``the performance of [IR]
systems is insensitive to ambiguity but very sensitive to erroneous
disambiguation'' (p 149).

\namecite{Schutze:97} first distinguishes sense {\em discrimination} from
disambiguation.  Discrimination involves identifying distinct
senses and classifying occurrences of the word as belonging to one of
those senses.  It does not involve labelling the senses (which
correspond to clusters of occurrences) or associating them with any
external knowledge source such as a dictionary.  Thus, in keeping with the spirit
of this paper, his senses are
automatically devised to match the corpus. 
System performance improved by up to
4.3\%.\footnote{They cite an improved average precision (over 11
levels of recall) of 14.4\% compared to the baseline, from 29.9\% to
34.2\%.  This improvement is 4.3\% in absolute terms, but 14.4\% when
calculated as an improvement on the baseline performance.}  with the
addition of the disambiguation module (and the added sophistication
that a word can be assigned to more than one word sense, where it is
`near' more than one in vector space).

It is debatable how important an improvement of 2 or 4 percentage
points is.  On the one hand, WSD will clearly not revolutionise IR or
render it a solved problem.
But IR is a fairly mature technology, very widely used by millions of
users, and an average 4\% improvement across all those users and all
their many queries could be seen as very significant indeed.

\subsection{Machine Translation}

In IR,  it is generally difficult to assign blame for
poor performance to word sense ambiguity or any other specific
source. MT, by contrast, wears its mistakes on its sleeve.  It is abundantly
clear to all in MT  that word sense ambiguity is a huge problem.

The literature has surprisingly little to say about it.
\namecite{HutchinsSomers:92} point out the two variants of the problem:
monolingual ambiguity, where the word is ambiguous in the source
language, and translational ambiguity, where speakers of the source
language do not consider the word ambiguous but it has two possible
translations, as when English {\em blue} is translated differently
into Russian according to whether it is light blue or dark.

MT is a technology rather than a science.  MT systems generally take
a decade from idea to marketplace, so the theory available at their inception is destined to
be out of date by the time they perform.  Thus no recent WSD work
is employed in existing MT systems.  They use
extensive sets of selection restrictions paired
with semantic features to make it possible for the system to make the
correct lexical choice.  MT systems usually use a number of
very large lexicons where selection restriction information, designed
to resolve ambiguity problems, accounts for a large proportion of the
bulk.  The SYSTRAN English-French lexicon responsible for word choice
contains 400 rules governing the one English word, {\em oil}, and when it
should be translated as {\em huile}, when {\em p\'{e}trole} \cite[p 179]{HutchinsSomers:92}.

One paper which does bring state-of-the-art WSD to bear on Machine
Translation, albeit in experimental mode, is \namecite{DaganItai:94}.
They use a bilingual lexicon to identify the possible translations,
and a  parsed target language corpus to gather information about the
`tuples' in which each of the possible translations is often
found.  A `tuple' comprises a grammatical relation, such as {\sc
subject-verb}, and the occupier of each of the slots of that relation,
so ``The man walked home'' would give the triple ({\sc
subject-verb}, man, walk).  The source-language text to be translated
in then parsed, to give a source language tuple.  The bilingual
dictionary and the target-language statistics are then
used to find the best match.

The paper applies sophisticated WSD to a real
problem, with the discriminations that the system makes
being defined by the needs of the application.

\subsection{Parsing}

Accurate parsing is a requirement for a wide range of NLP
applications, so if WSD is critical for parsing accurately, it is, by
implication, significant for all those applications that depend on parsing.
\namecite{McCarthy:97} explores WSD methods explicitly for purposes of
improving parsing.  Before assessing whether WS ambiguity is critical, let
us take a step back.  

It is well-established that ``the problem of
syntactic ambiguity is AI-complete'' \cite[p 269]{Hobbs:92}.  Here,
let us focus on one particular, but pervasive, variety of syntactic
ambiguity: prepositional phrase (PP) attachment.  A
problem is AI-complete if its solution requires a solution to all the
general AI problems of representing and reasoning about arbitrary
real-world knowledge.  In principle, any item of general knowledge
might be the datum required to make a PP-attachment.  If that
is all that can be said, the outlook is bleak.
We would hope that, in practice, a small and tractable subset of
general knowledge will resolve a high proportion of ambiguities.  

Some approaches to high-quality parsing make extensive use of
machine-readable dictionaries (MRDs).  In the 1990s, Microsoft have
been the leading proponents of `MRDs-for-parsing'.\footnote{The method
is used in the parser embedded in 1997 Microsoft Word's grammar
checker, as demonstrated by Steve Richardson at the ACL
Conference in Applied NLP, Washington D.C., 1997.} The hypothesis
behind the approach is that dictionary entries provide, implicitly or
explicitly, the information required to resolve most syntactic
ambiguities.

Note that, even if this hypothesis is true, it does not imply that WSD
has an important role to play.  Lexical information can resolve many
syntactic ambiguities without being sense-disambiguated.  
Consider 
\begin{description}
\item[1] I love baking cakes with friends.
\item[2] I love baking cakes with butter icing.
\end{description}
The PP attachment ambiguity is resolved, along with the ambiguity of
{\em with}, by the semantic class of the final noun phrase.  Where the
head of this noun phrase is human, as in 1, the PP attaches to the
verb.  Where it is a cake ingredient, it attaches to {\em cakes}.
Lexical information is required to determine the attachment in 1 and
2, but, since neither {\em friends} nor {\em icing} is ambiguous
between humans and cake-ingredients, disambiguation is not required.

That lexical
information will resolve a high proportion of syntactic ambiguities is
one hypothesis; that a significantly higher proportion will be
resolved, if the lexical information is sense-specific, is another.

Almost no work has been done to test either hypothesis.
\namecite{Whittemore:90} tested and confirmed a related hypothesis:  that `lexical
preferences' of nouns and verbs for PPs of a particular type are
better predictors of PP-attachment than any purely syntactic
considerations.  They took a sample corpus and counted the PPs that
would be correctly attached if each strategy was used.
To discover the significance of WS-ambiguity to parsing, a study
is required which combines this method with Krovetz and Croft's, of
manually disambiguating to determine the performance improvement that
would be achieved with a perfect WSD program.

\subsection{Lexicography}

NLP is most aware of lexicographers as suppliers of wares, but they
are also customers.  A linguistically annotated corpus is of more use
to a lexicographer than a `raw' one, as he or she can then investigate
the behaviour of a word in particular linguistic contexts without
having to trawl through large numbers of irrelevant citations.  A
sense-annotated corpus would be particularly valuable, as the
lexicographer would not have to trawl through `money {\em bank}'
citations when defining `river {\em bank}' \cite{Clear:94}.
There is then an intriguing possibility that the behaviour of WSD
programs will feed back into the nature of the dictionary senses they
disambiguate between.

\subsection{NLU}

For existing NLP applications requiring a deeper understanding of the text,
99\% of the ambiguity to be found in a desk dictionary is
not relevant.  This is, firstly, because these applications deal only
with very specific text types.  The specific sublanguage generally
means that, if a word has a meaning which is of interest, it is very
likely that occurrences of the word will be being used in that meaning
and not some other. Secondly, even then the application can only
interpret those inputs for which there is a possible interpretation in
the knowledge base (or in the system's output behaviour).  Several
respondents to the email survey, where I asked, ``does WS ambiguity
cause problems for your system?'',  commented ``We don't have any semantics in
our lexicon, we just have hooks into the knowledge representation''.

Where a word has one sense in the domain model, and one or more
outside it, an NLU application can generally determine whether the word is
being used in the domain sense by identifying whether the entire
sentence or query is coherent in terms of the domain model.  If it
is, the word is almost certainly being used in the domain sense. Where
a word has more than one domain sense, it is unlikely that both will
produce coherent analyses.  The domain model will generally provide
disambiguating material, not because it has been explicitly added, but
because type-checking and coherence-checking which is necessary in
any case will reject invalid senses.

With time, NLU systems will become more sophisticated, with richer
domain models and less limitations in the varieties of text they can
analyse.  This will make WSD more salient, though different strategies
will be relevant for the `foreground lexicon' containing the key words
for the domain model, and the `background lexicon', containing all
other words.  Foreground lexicon senses will be tightly-defined and
domain-specific, and will be disambiguated by coherence-checking.
Background lexicon disambiguation will only need to be between
coarse-grained senses.  Its function will be to increase parse
accuracy, and statistical methods will be appropriate. (The full
argument is presented in
\namecite{frascati}.)

\section{Answers}

The answers to the question, ``Does WS ambiguity cause problems
for NLP applications?'' are:

\begin{description}
\item[IR:] yes, to some moderate degree.  Problems can substantially
be overcome by using longer queries.  Within IR, WSD features as
something of an alternative to NLP.

\item[MT:] yes.  Huge problem, with the problem space defined by all
the one-to-many and many-to-many mappings in a bilingual
dictionary.  Addressed to date by lots and lots of selection
restrictions.

\item[Parsing:] not known.

\item[Lexicography:] yes, WSD would be of benefit.

\item[NLU:] not much.  NLU applications are mostly domain specific, and
have some sort of domain model.  It is generally necessary to have a
detailed knowledge of the word senses that are in the domain, so the
knowledge to disambiguate will often be available in the domain model
even where it has not explicitly been added for disambiguation
purposes.
\end{description}

\bibliographystyle{fullname}

\end{document}